\documentclass[sigconf]{acmart}
\AtBeginDocument{%
  }

\acmYear{2026}
\setcopyright{cc}
\setcctype{by-nc-nd}
\acmConference[WWW Companion '26]{Companion Proceedings of the ACM Web Conference 2026}{April 13--17, 2026}{Dubai, United Arab Emirates}
\acmBooktitle{Companion Proceedings of the ACM Web Conference 2026 (WWW Companion '26), April 13--17, 2026, Dubai, United Arab Emirates}
\acmPrice{}
\acmDOI{10.1145/3774905.3793118}
\acmISBN{979-8-4007-2308-7/2026/04}

\usepackage{listings}
\usepackage{balance} % for balancing columns on the final page
\usepackage{graphicx}
\usepackage{subfig}
\usepackage{multirow} 
\usepackage{makecell}
\usepackage{hyperref}

\usepackage{caption}

\begin{document}

%%
%% The "title" command has an optional parameter,
%% allowing the author to define a "short title" to be used in page headers.
\title{GCAgent: Enhancing Group Chat Communication through Dialogue Agents System}

%%
%% The "author" command and its associated commands are used to define
%% the authors and their affiliations.
%% Of note is the shared affiliation of the first two authors, and the
%% "authornote" and "authornotemark" commands
%% used to denote shared contribution to the research.
\author{Zijie Meng}
\authornote{Both authors contributed equally to this research.}
\affiliation{%
  \institution{Zhejiang University}
  \city{Hangzhou}
  \country{China}
}
\email{zijie.22@intl.zju.edu.cn}

\author{Zheyong Xie}
\authornotemark[1]
\affiliation{%
  \institution{Xiaohongshu Inc.}
  \city{Shanghai}
  \country{China}}
\email{xiezheyong@xiaohongshu.com}

\author{Zheyu Ye}
\affiliation{%
  \institution{Xiaohongshu Inc.}
  \city{Shanghai}
  \country{China}}
% \email{{{yezheyu, luchonggang}}}
\email{yezheyu@xiaohongshu.com}

\author{Chonggang Lu}
\affiliation{%
  \institution{Xiaohongshu Inc.}
  \city{Shanghai}
  \country{China}}
\email{luchonggang@xiaohongshu.com}

\author{Zuozhu Liu}
\affiliation{%
  \institution{Zhejiang University}
  \city{Hangzhou}
  \country{China}
}
\email{zuozhuliu@intl.zju.edu.cn}

\author{Zihan Niu}
\affiliation{%
  % \institution{University of Science and Technology of China, Hefei, China}
  \city{University of Science and Technology of China, Hefei}
  \country{China}
}
\email{niuzihan@mail.ustc.edu.cn}

\author{Yao Hu}
\affiliation{%
  \institution{Xiaohongshu Inc.}
  \city{Shanghai}
  \country{China}}
% \email{{{xiahou, caoshaosheng}}}
\email{xiahou@xiaohongshu.com}

\author{Shaosheng Cao}
\authornote{Corresponding author.}
\affiliation{%
  \institution{Xiaohongshu Inc.}
  \city{Shanghai}
  \country{China}}
% \email{{{xiahou, caoshaosheng}}}
\email{caoshaosheng@xiaohongshu.com}
%%
%% By default, the full list of authors will be used in the page
%% headers. Often, this list is too long, and will overlap
%% other information printed in the page headers. This command allows
%% the author to define a more concise list
%% of authors' names for this purpose.
\renewcommand{\shortauthors}{Zijie Meng et al.}

%%
%% The abstract is a short summary of the work to be presented in the
%% article.
\begin{abstract}
    As a key form in online social platforms, group chat is a popular space for interest exchange or problem-solving, but its effectiveness is often hindered by inactivity and management challenges. While recent large language models (LLMs) have powered impressive one-to-one conversational agents, their seamlessly integration into multi-participant conversations remains unexplored. To address this gap, we introduce GCAgent, an LLM-driven system for enhancing group chats communication with both entertainment- and utility-oriented dialogue agents. The system comprises three tightly integrated modules: Agent Builder, which customizes agents to align with users' interests; Dialogue Manager, which coordinates dialogue states and manage agent invocations; and Interface Plugins, which reduce interaction barriers by three distinct tools. Through extensive experiment, GCAgent achieved an average score of 4.68 across various criteria and was preferred in 51.04\% of cases compared to its base model. Additionally, in real-world deployments over 350 days, it increased message volume by 28.80\%, significantly improving group activity and engagement. Overall, this work presents a practical blueprint for extending LLM-based dialogue agent from one-party chats to multi-party group scenarios.
\end{abstract}

%%
%% The code below is generated by the tool at http://dl.acm.org/ccs.cfm.
%% Please copy and paste the code instead of the example below.
%%


%%
%% Keywords. The author(s) should pick words that accurately describe
%% the work being presented. Separate the keywords with commas.
%% A "teaser" image appears between the author and affiliation
%% information and the body of the document, and typically spans the
%% page.
% \begin{teaserfigure}
%   \includegraphics[width=\textwidth]{sampleteaser}
%   \caption{Seattle Mariners at Spring Training, 2010.}
%   \Description{Enjoying the baseball game from the third-base
%   seats. Ichiro Suzuki preparing to bat.}
%   \label{fig:teaser}
% \end{teaserfigure}

% \received{20 February 2007}
% \received[revised]{12 March 2009}
% \received[accepted]{5 June 2009}

%%
%% This command processes the author and affiliation and title
%% information and builds the first part of the formatted document.
\maketitle

\section{Introduction}
With the development of digital communication platforms, group chat has attracted significant attention due to its distinctive multi-participant interactive nature~\cite{warran2023online, yan2024people}. Unlike traditional one-to-one communication, group chat facilitates diverse conversational content, ranging from interactions among acquaintances to communications with strangers. However, the lack of fresh and engaging content from partially or entirely silent members often leads to inactivity within the group chat~\cite{popowski2024commit}. Additionally, the complex composition of members results in management difficulties, significantly impacting the effectiveness of group chat as a platform for interest exchange or problem-solving~\cite{nematzadeh2019information, ren2016impact}.

As the rapid advancement of Large Language Models (LLMs)~\cite{hurst2024gpt, qwen2}, the emergence of numerous dialogue agents built upon these models presents promising opportunities. However, mainstream AI agents from social platorms (such as Glow, Character AI, and My AI)\footnote{Glow: \url{https://glowconnect.org.uk}, Character AI: \url{https://character.ai}, My AI: \url{https://my.ai.se}.} and academic community~\cite{brandtzaeg2022my, niu2025part} are still predominantly limited to two-party dialogue scenarios. Moreover, GIFT~\cite{gu2023gift} injects conversational graph edges into attention for multi-party understanding but with limited performance. MUCA~\cite{mao2024muca} brings LLM into group chats to decide what to say, when to respond and whom to answer, but without exploring post-training and deployment. Therefore, effectively integrating dialogue agents into real-world group chat to enhance both content generation and operational assistance remains underexplored.

\begin{figure}[t]
    \centering
    \includegraphics[width=\linewidth]{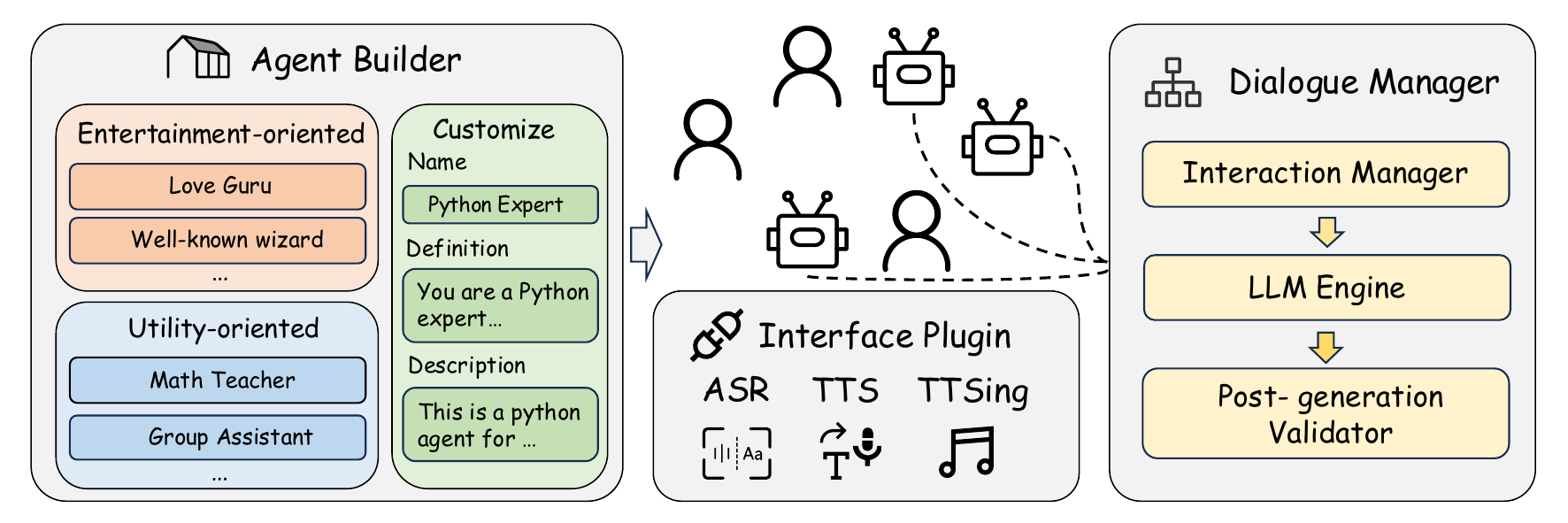}
    \caption{The overview of our GCAgent system.}
    \label{fig:enter-label}
\end{figure}

To address this gap, we propose \textbf{GCAgent} to enhance group chat communication through dialogue agents system, with a focus on delivering engaging content and supporting daily management, which is also defined as entertainment-oriented and utility-oriented. GCAgent system comprises three core components: \textbf{Agent Builder}, \textbf{Dialogue Manager}, and \textbf{Interface Plugins}. Specifically, the Agent Builder customizes agents according to personal interests. The Dialogue Manager coordinates the multi-party dialogue processes within group chats. Finally, the Interface Plugins facilitate smoother interaction through tools such as Automatic Speech Recognition (ASR)~\cite{asr}, Text-to-Speech (TTS)~\cite{tts} and Text-to-Sing (TTSing)~\cite{ttsing}, offering diverse communication modes to enhance user experience.

Through both offline and online evaluations, GCAgent achieved an average score of 4.68 across various evaluation criteria and a win rate of 51.04\% compared to its base model. It has also demonstrated consistent improvements in group activity, new group creation, message readership and message volumes, especially increases 28.80\% for message volumes. Furthermore, it has been deployed in real-world environments for over 350 days, providing exceptional service to numerous users. We also delivered a live demonstration on YouTube\footnote{\url{https://www.youtube.com/shorts/dsbQtNMqecc}}. 
Our contributions can be summarized as follows:

\begin{itemize}
    \item We developed GCAgent encompassing Agent Builder, Dialogue Manager, and Interface Plugins, to enhance group chat communication.
    \item Through extensive evaluation, we demonstrate its effectiveness across various dimensions.
    \item GCAgent has been seamlessly integrated into real-world group chat environments over 350 days, providing users an improved conversational experiences.
\end{itemize}

\begin{figure*}[t]
    \centering
    \subfloat[Agent Builder]{
       \label{fig:subfig:a}
       \includegraphics[width=0.12\textwidth]{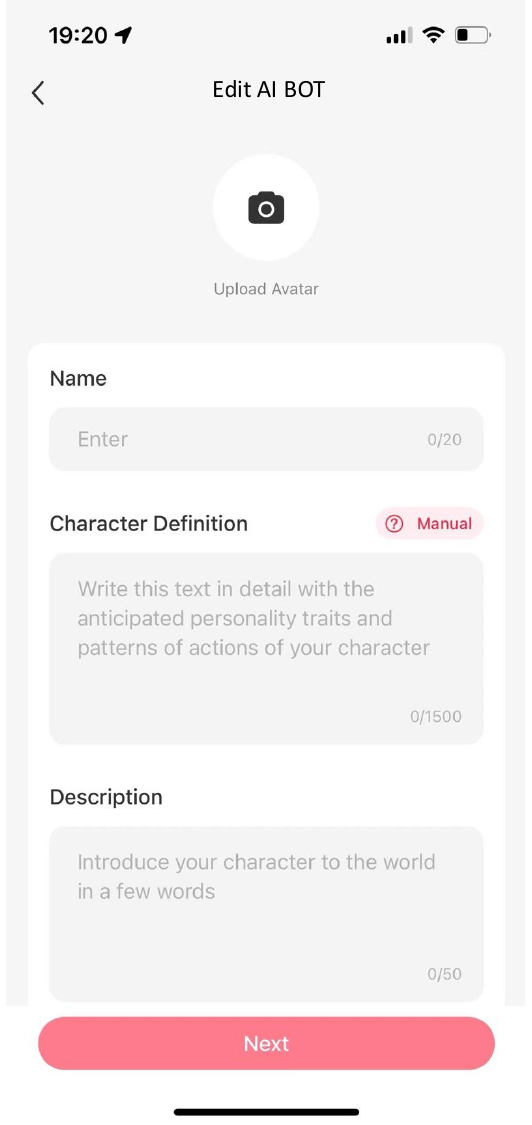}}
    \hspace{8mm}
    \subfloat[Voice Square]{
       \label{fig:subfig:b}
       \includegraphics[width=0.12\textwidth]{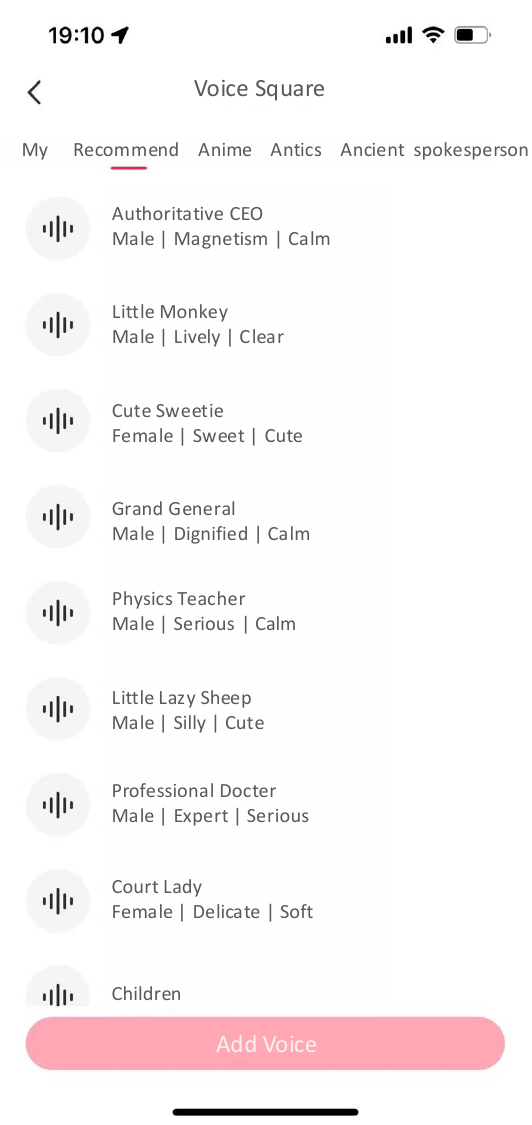}}
    \hspace{8mm}
    \subfloat[Agent Square]{
       \label{fig:subfig:c}
       \includegraphics[width=0.12\textwidth]{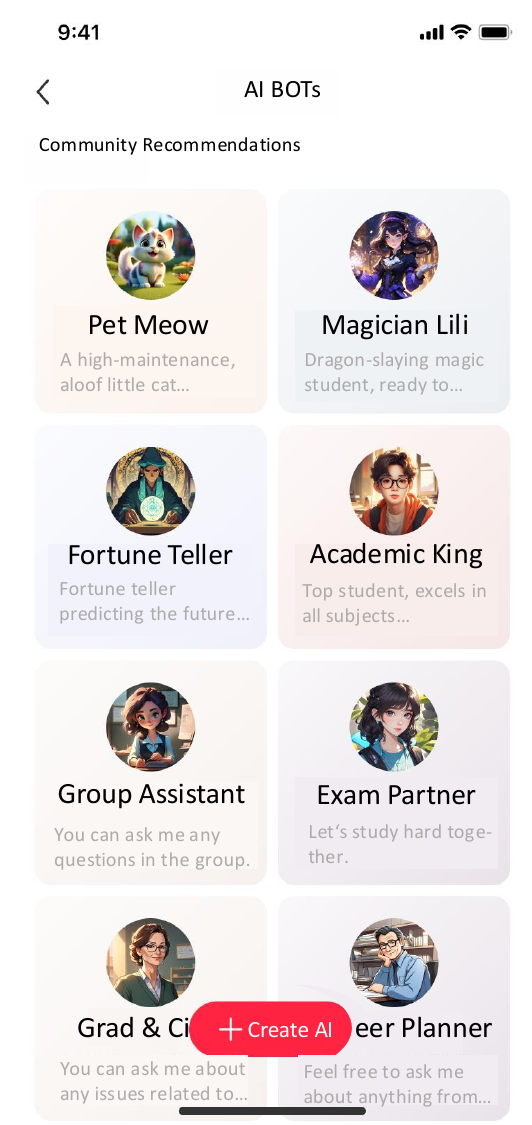}}
    \hspace{8mm}
    \subfloat[Agent Usage]{
       \label{fig:subfig:d}
       \includegraphics[width=0.12\textwidth]{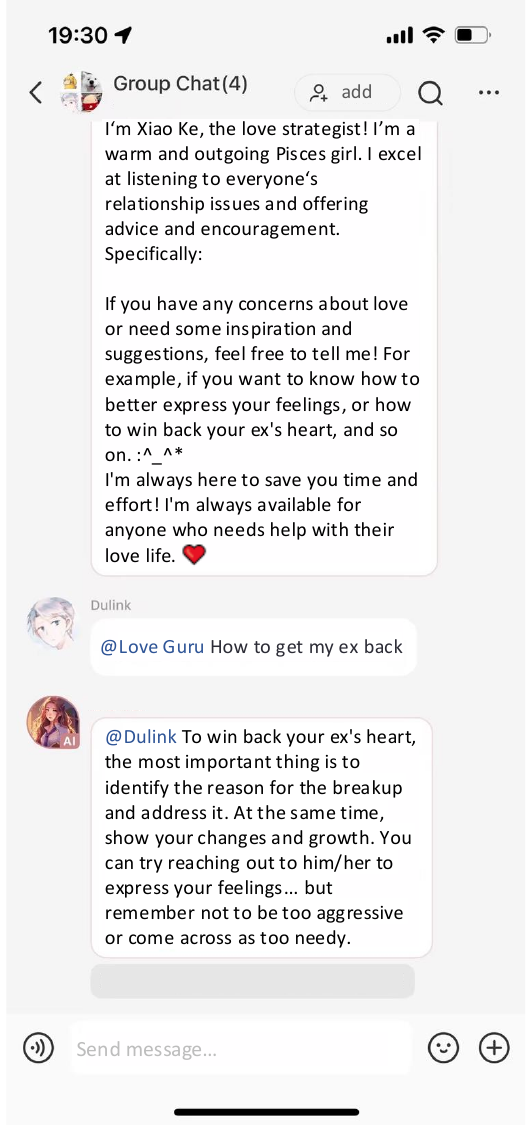}}
    \hspace{8mm}
    \subfloat[Plugin Usage]{
       \label{fig:subfig:e}
       \includegraphics[width=0.12\textwidth]{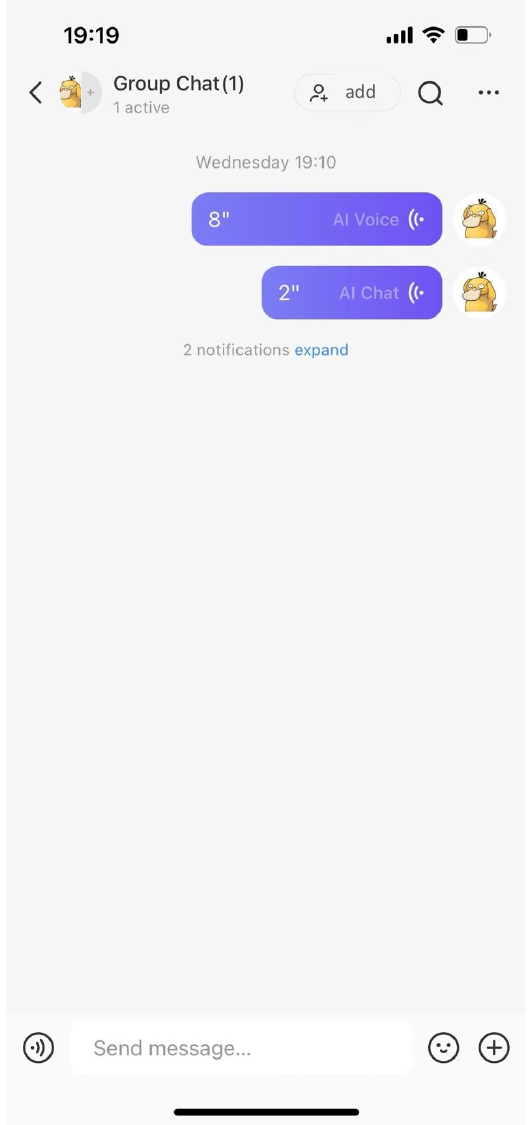}}
    \caption{The user interface of GCAgent. And a brief live demonstration is available on \href{https://www.youtube.com/shorts/dsbQtNMqecc}{YouTube}.}
    \label{fig:interface}
\end{figure*}

\section{Design and Implementation of System}
\subsection{Agent Builder}
The Agent Builder is a powerful tool designed for customizing and creating group chat agent. As shown in Figures~\ref{fig:subfig:a} and~\ref{fig:subfig:b}, users can easily design agents by filling in the required fields and selecting a preferred voice style. Additionally, we provide a range of predefined agents, categorized into two primary types: entertainment-oriented and utility-oriented, as shown in Figure~\ref{fig:subfig:c}. They encompass a broad spectrum of personality traits, thereby enhancing community engagement, fostering emotional connections, improving group interactions and addressing specific demands.

\subsection{Dialogue Manager}
The Dialogue Manager is composed of the Interaction Manager to manage dialogue states and agent invocations, the LLM Engine to generate natural, context-aware responses using LLMs, and the Post-generation Validator to ensure the quality and relevance of responses through automated checks and error corrections. They work together to ensure the stability of agent dialogue management and the high quality of generated responses.

\textbf{Interaction Manager} coordinates dialogue collection, agent invocation, and information recording to ensure coherent and personalized responses in group chat communication. Specifically, it tracks complex conversation threads by incorporating historical dialogue records, user behavior, and relevant contextual information. In multi-parity scenarios, as shown in Figure \ref{fig:subfig:d}, users can invoke specific agents by ``\textbf{@}'' tag, prompting their participation in group chat sessions. Additionally, the Interaction Manager monitors message sequencing, participant management, and session tracking to ensure orderly and effective interactions across all chat scenarios.

\textbf{LLM Engine} is the core component for understanding and generating responses. Our developed model builds upon Qwen2-7B-Instruct~\cite{qwen2} through fine-tuning using a vast corpus of real dialogue data, resulting in enhanced conversational abilities and better contextual adaptation.

\textbf{Post-generation Validator} ensures the agent's generated responses quality through automated checks. It corrects grammatical and semantic errors using advanced regular expressions and evaluation methods, and implements a retry mechanism to regenerate responses that fail to meet quality standards, ultimately enhancing content reliability and user satisfaction.

\subsection{Interface Plugins}
Plugins serve as essential tools that complement these agents by enabling specific functionalities and improving conversational transitions. We provide three distinct plugins: ASR~\cite{asr}, TTS~\cite{tts}, and TTSing~\cite{ttsing}. The first two plugins facilitate seamless communication between users and dialogue agents through bidirectional conversion between speech and text. Meanwhile, TTSing provides both users and agents with the ability to convert text into songs, addressing entertainment demands during group conversations and significantly enhancing the overall user experience.

\section{Experiments}
\subsection{Offline Evaluation}
\subsubsection{Experimental Setting}
% train: 33569  test: 3000
In the offline experiments, we compared GCAgent with its base model, Qwen2-7B-Instruct (Qwen)~\cite{qwen2}. Specifically, we curated 36,569 anonymized group‐chat samples, using 3,000 for testing and the remainder for fine-tuning. Each entry includes the role configuration of GCAgent, historical conversations, the most recent user message, and the corresponding LLM-generated responses, all of which were manually validated. Then, we devised two methodologies: direct scoring and indirect comparison, and we adopted a GPT4o-based~\cite{hurst2024gpt} LLM-as-a-judge framework, which has a high correlation with human judgments~\cite{gu2024survey, chiang2023closer}, to balance quality and efficiency. In the direct scoring, we used four commonly adopted criteria: 1) Correctness~\cite{she-etal-2024-exploring, xue-etal-2023-improving}: assessing whether the model accurately comprehends user intent and provides correct solutions or information in its responses. 2) Consistency~\cite{li2024dialogues}: evaluating whether the model’s responses align with its predefined role and context, maintaining conversational coherence and logical flow in multi-turn dialogues. 3) Fairness~\cite{zhao-etal-2023-chbias, lin2023toxicchat}: determining whether the model generates unbiased, non-discriminatory, and ethically appropriate content, avoiding fabricated or inappropriate material. 4) Engagement~\cite{ghazarian2020predictive, xu-etal-2022-endex}: measuring whether the responses are readable, comprehensible, concise, and emotionally satisfying. Each dimension was scored discretely from 1 to 5, with 1 for poor, 2 for fair, 3 for moderate, 4 for good, and 5 for excellent. In the indirect comparison, these four criteria were used to guide GPT4o~\cite{hurst2024gpt} to identify a winner or declare a tie between two candidate responses. To mitigate position bias~\cite{pezeshkpour2023large}, each pair was evaluated twice in reversed order, with contradictory outcomes marked as ties. Additionally, to enhance the evaluation accuracy, we implemented an ``analyze-rate'' method~\cite{chiang2023closer}, prompting the LLM to present a detailed rationale before final scoring. 

\begin{table}[t]
    \centering
    \small
    \caption{Results of direct scoring in offline evaluation.}
    \label{tab:direct_scoring}
    \resizebox{\linewidth}{!}{%
        \begin{tabular}{c|ccccc}
            \toprule
            \multirow{2}{*}{Models} & \multicolumn{5}{c}{Direct scoring} \\
            % \cmidrule{2-6}
             & Correctness & Consistency & Fairness & Engagement & Average \\
            \midrule
            Qwen & 4.18 & 4.33 & 4.90 & 4.27 & 4.42 \\
            GCAgent & \textbf{4.40} & \textbf{4.79} & \textbf{4.94} & \textbf{4.59} & \textbf{4.68} \\
            \bottomrule
        \end{tabular}%
    }
\end{table}

\begin{table}[t]
    \centering
    \small
    \caption{Results of indirect comparison in offline evaluation.}
    \label{tab:indirect_cmp}
    \begin{tabular}{c|ccc}
        \toprule
        \multirow{2.5}{*}{\makecell[c]{GCAgent\\vs Qwen}} & Win & Tie & Lose \\
        \cmidrule{2-4}
         & 51.04\% & 29.57\% & 19.39\% \\
        \bottomrule
    \end{tabular}
\end{table}

\subsubsection{Experimental Result}
As shown in Table~\ref{tab:direct_scoring} and Table~\ref{tab:indirect_cmp}, we compared GCAgent and its base model (Qwen) from multiple perspectives. In the direct scoring, GCAgent outperformed Qwen by an average of 0.26 across four criteria, demonstrating superior adaptability to group chat scenarios. And GCAgent achieved 4.94 in fairness, indicating strong adherence to community guidelines and a low likelihood of generating inappropriate content. GCAgent also surpassed the base model in consistency by 0.46, reflecting a more nuanced understanding of role definitions and conversational context, which further contributes to higher scores in correctness and engagement. In the indirect comparison, Qwen was preferred in only 19.39\% of the entries, whereas GCAgent performed better in over half. Overall, these experiments demonstrate that GCAgent can effectively maintain a healthy and safe chat environment while fostering deeper user interactions, enhancing group activity.

\subsection{Online Evaluation}
\begin{table}[t]
    \centering
    \small
    \caption{Online evaluation results. Unique user views is used to assess message readership, and amount to others.}
    \label{tab:online_eval}
    \resizebox{\linewidth}{!}{%
        \begin{tabular}{c|cccc}
            \toprule
            Metric & \makecell{Group\\Activity} & \makecell{New Group\\Creation} & \makecell{Message\\Readership} & \makecell{Message\\Volumes} \\
            \midrule
            Improvement (\%) & +4.02 & +6.27 & +11.07 & +28.80 \\
            \bottomrule
        \end{tabular}%
    }
\end{table}

We deployed GCAgent across numerous chat groups and conducted A/B test. As shown in Table~\ref{tab:online_eval}, integrating agents into group chats significantly enhanced activity. Specifically, it increased the proportion of groups activity by 4.02\%, newly group creation by 6.27\%, message readership by 11.07\%, and the messages volumes role by 28.80\%. In terms of user retention, over 12\% of users initiated conversations with GCAgents when they joined an agent-enabled group. And the retention rates exceeded 30\% for next-day, 15\% for three-day, and 10\% for seven-day. Additionally, weekly active users consisted of 35\% adults and 65\% minors, with minors nearly twice as active as adults on weekends and holidays.

\begin{figure*}[h]
    \centering
    \includegraphics[width=.79\textwidth]{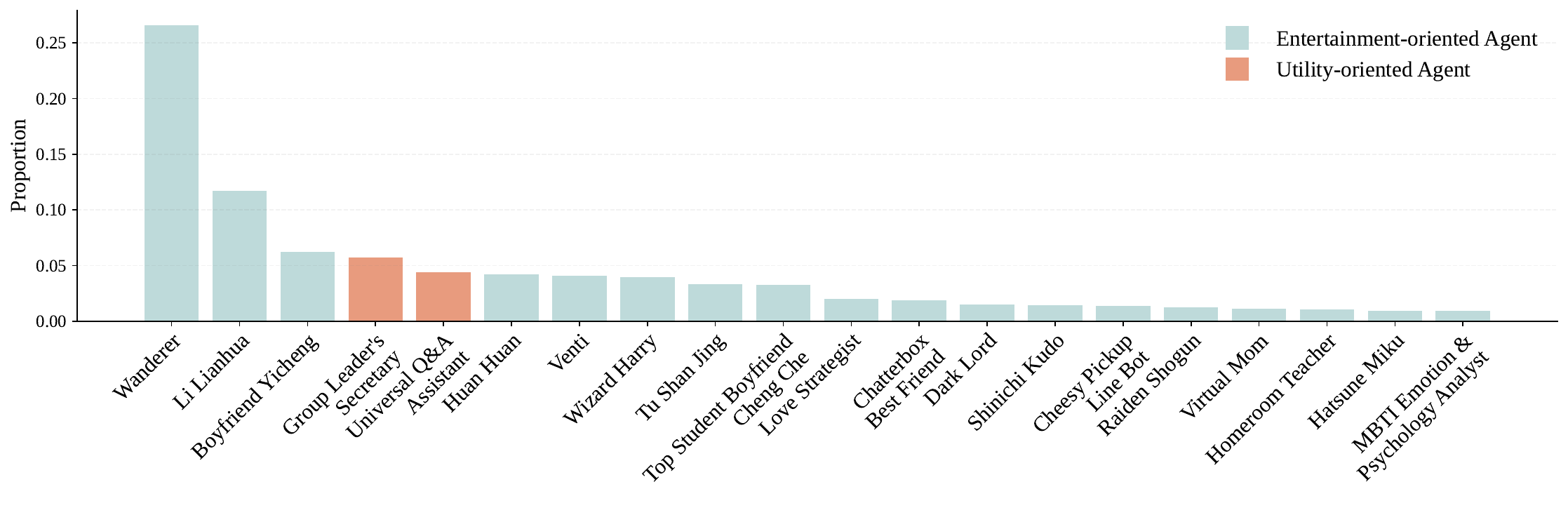}
    \caption{Role distribution of top 20 most popular agents.}
    \label{fig:role_distri}
\end{figure*}
\subsection{User Analysis}
The current system has generated over one million agents, of which 97\% are entertainment-oriented agents designed for emotional companionship and casual conversation. In contrast, only 3\% are utility-oriented agents, typically serving as group assistants or domain-specific problem solvers. Analysis of user interaction data reveals that entertainment-oriented agents participate in an average of 18 conversations per day and maintain a next-day retention rate of 25\%. In highly active, entertainment-focused group chats, more than 10\% of conversations involve agents, highlighting their central role in social interaction. Conversely, utility-oriented agents exhibit significantly lower engagement levels. As illustrated in Figure~\ref{fig:role_distri}, among the top 20 most popular agents, 18 are entertainment-oriented, with only two—``Group Leader's Secretary'' and ``Universal Q\&A Expert''—falling into the utility-oriented category. These two agents respond with an average of only 3 messages per user and achieve a modest 9\% next-day retention rate,  which is substantially lower than their entertainment-oriented counterparts.

\section{Conclusion and Future Work}
GCAgent demonstrates the feasibility of deploying LLM based agents into group chat scenarios. By integrating the Agent Builder, Dialogue Manager, and Interface Plugins, it increased message volume by 28.80\% and maintained high user satisfaction over a 350-day deployment, revitalizing dormant groups into active spaces for interest exchange and problem-solving. In the future, we will extend it to multilingual, multimodal, cross-platform environments, incorporating enhanced safety, provenance, and consent mechanisms, along with vision- and document-aware plugins that enable agents to ground their responses in shared media.

\section{Acknowledgement}
This work is partially supported by the National Key R\&D Program of China (Grant No. 2024YFC3308304), the "Pioneer" and "Leading Goose" R\&D Program of Zhejiang (Grant no. 2025C01128).

%%
%% The next two lines define the bibliography style to be used, and
%% the bibliography file.
\bibliographystyle{ACM-Reference-Format}
\balance
\bibliography{sample-base}

%%
%% If your work has an appendix, this is the place to put it.
% \appendix

% \section{Research Methods}

% \subsection{Part One}

% Lorem ipsum dolor sit amet, consectetur adipiscing elit. Morbi
% malesuada, quam in pulvinar varius, metus nunc fermentum urna, id
% sollicitudin purus odio sit amet enim. Aliquam ullamcorper eu ipsum
% vel mollis. Curabitur quis dictum nisl. Phasellus vel semper risus, et
% lacinia dolor. Integer ultricies commodo sem nec semper.

% \subsection{Part Two}

% Etiam commodo feugiat nisl pulvinar pellentesque. Etiam auctor sodales
% ligula, non varius nibh pulvinar semper. Suspendisse nec lectus non
% ipsum convallis congue hendrerit vitae sapien. Donec at laoreet
% eros. Vivamus non purus placerat, scelerisque diam eu, cursus
% ante. Etiam aliquam tortor auctor efficitur mattis.

% \section{Online Resources}

% Nam id fermentum dui. Suspendisse sagittis tortor a nulla mollis, in
% pulvinar ex pretium. Sed interdum orci quis metus euismod, et sagittis
% enim maximus. Vestibulum gravida massa ut felis suscipit
% congue. Quisque mattis elit a risus ultrices commodo venenatis eget
% dui. Etiam sagittis eleifend elementum.

% Nam interdum magna at lectus dignissim, ac dignissim lorem
% rhoncus. Maecenas eu arcu ac neque placerat aliquam. Nunc pulvinar
% massa et mattis lacinia.

\end{document}